\documentclass[
]{ceurart}

\usepackage{listings}
\begin{document}

\copyrightyear{2022}
\copyrightclause{Copyright for this paper by its authors.
  Use permitted under Creative Commons License Attribution 4.0
  International (CC BY 4.0).}

\conference{CLEF 2022: Conference and Labs of the Evaluation Forum, 
    September 5--8, 2022, Bologna, Italy}


\title{SINAI at eRisk@CLEF 2022: Approaching Early Detection of Gambling and Eating Disorders with Natural Language Processing}


\author{Alba María Mármol-Romero}[%
orcid=0000-0001-7952-4541,
email=amarmol@ujaen.es,
]
\author{Salud María Jiménez-Zafra}[%
orcid=0000-0003-3274-8825,email=sjzafra@ujaen.es
]
\cormark[1]

\author{Flor Miriam Plaza-del-Arco}[%
orcid=0000-0002-3020-5512,
email=fmplaza@ujaen.es,
]
\author{M. Dolores Molina-González}[%
orcid=0000-0002-8348-7154,
email=mdmolina@ujaen.es,
]
\author{María-Teresa Martín-Valdivia}[%
orcid=0000-0002-2874-0401,
email=maite@ujaen.es,
]
\author{Arturo Montejo-Ráez}[%
orcid=0000-0002-8643-2714,
email=amontejo@ujaen.es,
]
\address{Computer Science Department, SINAI, CEATIC, Universidad de Jaén, 23071, Spain}


\begin{abstract}
  This paper describes the participation of the SINAI team in the eRisk@CLEF lab. Specifically, two of the proposed tasks have been addressed: i) Task 1 on the early detection of signs of pathological gambling, and ii) Task 3 on measuring the severity of the signs of eating disorders. The approach presented in Task 1 is based on the use of sentence embeddings from Transformers with features related to volumetry, lexical diversity, complexity metrics, and emotion related scores, while the approach for Task 3 is based on text similarity estimation using contextualized word embeddings from Transformers. In Task 1, our team has been ranked in second position, with an F1 score of 0.808, out of 41 participant submissions. In Task 3, our team also placed second out of a total of 3 participating teams.
  
\end{abstract}


\maketitle

\section{Introduction}

The large amount of content posted daily on social media has made them a significant source of data for early detection of mental disorders and risky behaviours. The eRisk@CLEF 2022 lab \cite{erisk2022} focuses on early risk prediction on the Internet and its goal is to promote the development of automatic systems for the detection of mental disorders such as depression, self-harm or eating disorders. In this edition, three tasks have been proposed:
\begin{itemize}
    \item Task 1: Early Detection of Signs of Pathological Gambling. It involves sequentially processing writings and detecting as early as possible the first signs of pathological gambling. It is a continuation of the Task 1 proposed for eRisk 2021, but the difference is that in this edition training data has been provided, whereas last year it was an ``only test'' task. The training data of this edition comprises all test users of the 2021 task.
    \item Task 2: Early Detection of Depression. It consists of sequentially processing pieces of evidence and detect early traces of depression as soon as possible. It is a continuation of eRisk 2017 pilot task and eRisk 2018 Task 1. The difference between this edition and past editions is in the training set provided. In 2022, the training data is composed of all 2017 pilot task users (training users + test users) plus 2018 Task 1 test users.
    \item Task 3: Measuring the severity of the signs of Eating Disorders. Its aim is to estimate a user's level of disordered eating from his or her history of posts. For this purpose, for each user, a standard eating disorder questionnaire (EDE-Q) has to be filled in. This task is new in this edition and no training data are provided, that is, it is an ``only test'' task.
\end{itemize}

Currently, our research group SINAI\footnote{\url{https://sinai.ujaen.es}} is working on the Big Hug project\footnote{\url{https://bighug.ujaen.es}} focused on the early detection of disorders and misbehaviors (depression, anxiety, eating disorders, gambling addiction, suicidal ideation and cyberbullying)  in online social networks. Therefore, our interest in developing systems as those expected to answer eRisk tasks is high, as it is a perfect playground to test our approaches.

In this sense, our main goals are not only to produce systems reporting high performance but to understand the best methods and approaches that can be applied in similar scenarios. It is not our aim to put as many features and as many systems as possible all together in an ensemble of predictors to gain the top ranking position, but rather to find out the best approaches that can be applied to our project's objectives. The design of online and monitoring tools, as is requested in Task 1, along with the ability to understand user's disorder, the main pursuit in Task 3, fully matches our research interests.

This work presents the participation of our research group, the SINAI team, in \textit{Task 1: Early Detection of Signs of Pathological Gambling} and \textit{Task 3: Measuring the severity of the signs of Eating Disorders}.
The rest of the paper is organized as follows. Sections 2 and 3 describe the details of our participation in Task 1 and Task 3, respectively. Each of them is divided into subsections in which, first, we introduce what the task consists of, the data provided and the evaluation measures used. Secondly, the system developed and the methodology used are presented. Thirdly, the experimental setup is detailed. Subsequently, the results obtained and a discussion of them are presented. Finally, Section 4 shows the conclusions obtained after participation in the eRisk lab and the perspectives for future work.

\section{Task 1: Early Detection of Signs of Pathological Gambling}

\subsection{Task description}

This task focuses on early risk detection of gambling addiction by processing posts from social media in strict order of publication. The participant systems had to read the posts (from several users) in the order in which they were created, process them and generate a response in order to get the next posts. The data is composed of 14,627 posts by 81 different subjects categorized as positive in gambling addiction, and about 1M posts by 1,998 subjects not categorized as addicts \cite{erisk2022}. 

The task is faced from two different perspectives: as a binary decision problem, and as a ranking (regression) decision problem. As a binary decision problem, posts have to be labelled as positive (label 1, i.e. addiction detected) or negative (label 0, no addiction detected). The earlier the system detects an addiction, the better, as it is reflected with the ERDE and $F_{latency}$ metrics proposed by the organizers and used to evaluate the systems, along with the well known precision, recall and F1 scores. As a ranking decision problem, instead of assigning 0 or 1 labels, a score of the estimation of the risk to suffer such a disorder is computed. Different metrics as the ones used for information retrieval are considered to evaluate this second view of the task (P@10 or NDCG, among others).

\subsection{System and methods}

In order to address this task, we have followed a supervised learning approach. To train our models, we have used the training dataset provided by the eRisk organizers. This dataset consists of a time series of posts published by different users. Due to the limited sequence length of transformer models used (which will be described later), we made numerous tests to choose the right number of posts that would be more representative of the pathology (from the oldest ones to the newest ones). Finally, the 50 most recent posts were taken after evaluating different sizes.

An important part on which prediction models are based is the extraction of a feature vector from each set of posts per user provided by the organization. The features extracted by our system collect different aspects: volumetry, lexical diversity, complexity metrics, and emotion related scores. We explain in more detail what these features are about and the resources used to produce them.

\begin{itemize}
    \item Volumetry: we extract the number of words, number of unique words, number of characters, word average length, number of unique lemmas, long average of lemmas and number of each part-of-speech found. This was done with the Python package spaCy \cite{spacy2}.

    \item Lexical diversity: we apply different techniques to measuring lexical diversity, such as simple Type-Token Ratio (TTR), root TTR, log TTR, Maas TTR, Mean Segmental Type–Token Ratio (MSTTR), Moving-Average Type–Token Ratio (MATTR), Hypergeometric Distribution Diversity (HDD) and Measure of Textual Lexical Diversity (MTLD) \cite{mccarthy_mtld_2010}.

    \item Complexity: we apply different techniques to measuring the text complexity, such as the lexical complexity, Spaulding readability, sentence complexity, automated readability index, height of the dependency tree, punctuation marks, Fernández-Huerta's readability, Flesch-Szigrist readability, Gutierrez's comprehensibility, readability, minimum age of comprehension and SOL metric \cite{lopez-anguita_legibilidad_2018}.

    \item Emotions: this feature is focused on measuring the different emotions that are expressing in posts, such as fear, anger, joy or sadness. We obtain these emotions using two available pre-trained language models based on the Transformer architecture \cite{transformers}: DistilBERT\footnote{\url{https://huggingface.co/bhadresh-savani/distilbert-base-uncased-emotion}} \cite{Sanh2019DistilBERTAD} model fine-tuned on an annotated Twitter corpus on emotions \cite{saravia-etal-2018-carer} and a BERT\footnote{\url{https://huggingface.co/bhadresh-savani/bert-base-go-emotion}} \cite{kenton2019bert} model fine-tuned on an annotated Reddit corpus on emotions \cite{goemotions}.
\end{itemize}

\begin{figure}[!h]
        \centering
        \includegraphics[width=1\textwidth]{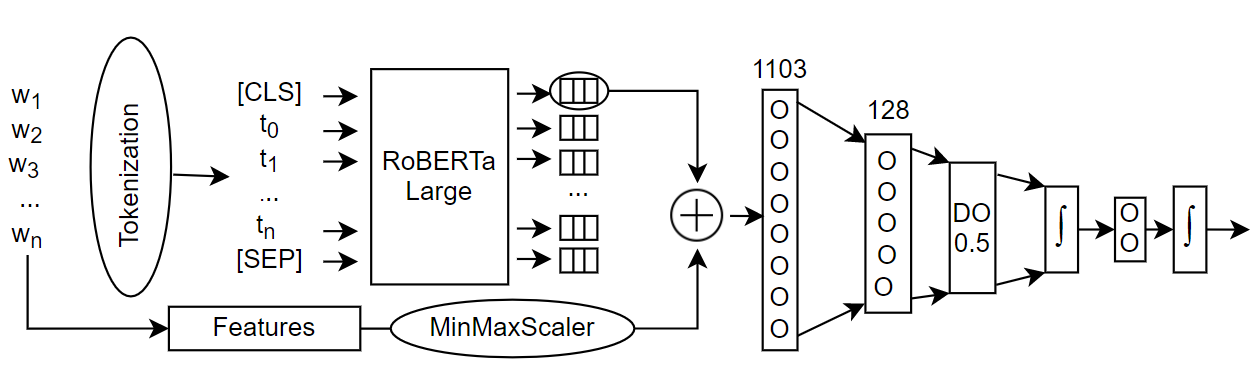}
        \caption{Model architecture for the binary decision task. The input representation is RoBERTa-Large tokenization. Each embedding of the model output is concatenated with its corresponding normalized feature vector. The difference for the classification task is the last FFN with 1 exit instead of 2 exits.}
        \label{model1}
    \end{figure}

    For our participation in eRisk Task 1 we have trained a total of three models, all supported by the RoBERTa-large \cite{roberta} linguistic model and feature vectors. Sentence embeddings from RoBERTa are concatenated with the features described above after scaling them into the [0, 1] interval, as can be seen in Figure \ref{model1}.
    
    As first step, the process we follow to predict task 1 consists in the extraction of the posts. The last 50 posts of each user are concatenated. This concatenation of posts is the input to the RoBERTa large automodel whose output is concatenated with the previously generated and normalized feature vectors (those about volumetry, complexity, lexical diversity and emotions). Once all features have been concatenated, a Feed Forward Neural Network (FFNN) with one hidden layer is applied to generate the final predictions. The first layer of the final FFN has an input dimension of 1103 (1024 from the embeddings vector generated by RoBERTa and 79 additional features). The output of the layer has a size of 128. These outputs are passed through a dropout layer (with probability 0.5) during learning and, finally, a ReLU activation function is also applied before feeding a last FFN with 128 inputs and 1 or 2 outputs (the final decision classes depending on the configuration of the run, as described below).
    
    The system was implemented using the Python packages scikit-learn\cite{scikit}, Transformers\cite{transformers} and PyTorch\cite{pytorch}, and trained on a 2xGPU NVIDIA V100 server.

\subsection{Experimental setup}

In the learning phase, a maximum length of 512 tokens per document was set. The document is the concatenation of the last 50 posts of a user (as described before). The learning rate for fine-tuning the neural network was set to 5e-5, the batch size was 8 and the number of train epochs was one. The optimizer used was AdamW. To explore the best parameters and evaluate the whole system, 5-fold cross validation approach was followed.

All runs follow the same structure. The runs differ in minor configuration aspects, but that triggered major changes in the prediction.

\begin{itemize}
    
    \item \textbf{Run 0}. This run consists of running a regression model in which the set of posts used has no pre-processing. The result of this model marks the score that a user has to be considered a potential gambler or not. Users with a score lower than 0.5 will not be considered potential gamblers, while those with a score equal to or higher than 0.5 will be considered potential gamblers.
    \item \textbf{Run 1}. In this run, the only difference with respect to run 0 is the processing of the concatenated posts. Before extracting the feature vector, the URLs found in the texts are removed (also brackets and parentheses containing URLs). 
    \item \textbf{Run 2}. This run is very similar to run 0. The set of texts used does not have any processing, however, the model built is a binary classification model (instead of a regression one), so the result of this model indicates the score that a user has to be considered and not considered a pathological gambler. The model generates a score for the two considered classes (addicted or not), which will define the final class after a softmax transformation.
    \end{itemize}

\subsection{Results and discussion}

Salient results have been achieved with the approaches explored by our team. From the reported results provided by the organizers, we have extracted our scores, which are shown in Tables \ref{tab:task1_results1} and \ref{tab:task1_results2}.

\begin{table}[htp!]
    \caption{Results of SINAI team for Task 1 in decision-based evaluation}
    \centering
    \begin{tabular}{ccccccccc}
         Run & $P$ & $R$ & $F1$ & $ERDE_5$ & $ERDE_{50}$ & $latency_{tp}$ & $speed$ & $latency_w F1$ \\ 
         \hline
         0 & 0.425 & 0.765 & 0.546 & 0.015 & 0.011 & 1.0 & 1.000 & 0.546 \\
         1 & 0.575 & 0.802 & 0.670 & 0.015 & 0.009 & 1.0 & 1.000 & 0.670 \\
         2 & 0.908 & 0.728 & 0.808 & 0.016 & 0.011 & 1.0 & 1.000 & 0.808 \\
         
    \end{tabular}
    \label{tab:task1_results1}
\end{table}

In decision-based evaluation, the value of F1 score obtained by our Run 2 (0.808) is the second highest among all submissions by participants (41 submissions were reported in total). Our ERDE values are the first and second highest (0.015 and 0.016). In terms of $speed$ and $latency_{TP}$ we also reach top values. The system that generated Run 2 is highly performant in this task, despite the limited amount of resources involved.

\begin{table}[!htp]
    \caption{Results of SINAI team for Task 1 in ranking-based evaluation (only 1 writing results reported)}
    \centering
    \begin{tabular}{cccc}
        Run & P@10 & NDCG@10 & NDCG@100 \\ \hline
        0 & 0.10 & 0.19 & 0.56 \\
        1 & 0.70 & 0.65 & 0.62 \\
        2 & 1.00 & 1.00 & 0.70 \\
    \end{tabular}

    \label{tab:task1_results2}
\end{table}

Regarding ranking-based evaluation (see Table~\ref{tab:task1_results2}), our team reaches a third position with values of 1.0 for both, P@10 and NDCG@10, and 0.7 for NDCG@100. It can be noticed that these scores are always zero for our system above 1 writing. We believe this is due to the limited number of submissions that our system needs to trigger an alert. Therefore, not enough scores are provided for a number of submissions above 100 for the ranking evaluation to be feasible. 

Our results indicate that the binary configuration is more convenient to train the system. The reason may be that, in the fist two configurations, determining the threshold after applying a sigmoid function on the output logit is not something the model can learn, as it is fixed on 0.5, while using softmax on two classes allows for a better exploration through the loss space. Additionally, further analysis is needed to determine how non-embeddings features contribute to the performance beyond pure end-to-end models. 

\section{Task 3: Measuring the severity of the signs of Eating Disorders}
\subsection{Task description}


In previous years, early detection of anorexia signs has been conducted in eRisk@CLEF \cite{losada2018overview,davide2019overview,Arco2019IntegratingUF}. This year, however, Task 3 ``Measure the severity of signs of eating disorders'' is proposed for the first time, which consists of an estimation of the level of characteristics associated with an eating disorder diagnosis from a user's writing history. For each user, a submission history is provided and participants are required to automatically fill in a standard eating disorder questionnaire. An important aspect is that no training data is provided to address this task.

The questionnaires are defined on the basis of the Eating Disorder Examination Questionnaire (EDE-Q). This questionnaire is designed to assess the range and severity of multiple characteristics associated with eating disorders. From it, the organizers have used only the questions 1-12 and 19-28. It employs four subscales: restraint (RS), preoccupation with eating (ECS), preoccupation with shape (CSC), preoccupation with weight (WCS), and a global score (GED). To obtain a particular subscale score, the ratings for the relevant questions (numbered in evaluation metrics) are added together and the sum divided by the total number of questions forming the subscale. To obtain an overall or ``global'' score, the four subscales scores are summed and the resulting total divided by the number of subscales (i.e. four). 












The official evaluation metrics of the competition are the following: MZOE, MAE, MAE$_{macro}$, GED, RS, ECS, SCS, WCS. Refer to the overview of the task \cite{erisk2022} for a detailed explanation of these measures.

\subsection{System and methods}

In this section, the methodology developed to address Task 3 is described.

First of all, we performed a preprocessing step to select those posts corresponding to the last 28 days of the user's history to follow the instructions given in the questionnaire: ``The following questions are concerned with the past four weeks (28 days) only''.

Once these posts were selected, the next step was to define the methodology used to answer the questionnaire. For this aim, we followed a method based on text similarity estimation using contextualized word embeddings from Transformers. Specifically, we computed the similarity between each post in a user's history with each question in the EDE-Q Questionnaire. This value ranges from 0 to 1 where a value close to 0 indicates that the post and the question are not similar while a value close to 1 indicates the opposite.

After computing this similarity, we adopted a heuristic to answer the questions given. For this aim, we first differentiated two types of questions that can be found in the eating disorder questionnaire: day-based questions and scale-based questions. The day-based questions are those whose answers are of the form: no days, 1-5 days, 6-12 days, etc., while the answers to the scale-based questions are as follows: not at all, slightly, moderately, etc.

On the one hand, to answer the day-based questions, we calculated the number of days that the user talks about the topic of the question. For this, the posts whose similarity with the question is greater than a threshold value (0.4 for run 1, 0.35 for run 2 and 0.375 for run 3) are selected. On the selected posts, the date of the first and the last post is chosen and the difference in days between the two is calculated. After that, we selected the option that matched the number of days obtained:

\begin{enumerate}
  \addtocounter{enumi}{-1}
  \item NO DAYS (0 days)
  \item 1-5 DAYS (1<= days <= 5)
  \item 6-12 DAYS (6 <= days <= 12)
  \item 13-15 DAYS (13 <= days <= 15)
  \item 16-22 DAYS (16 <= days <= 22)
  \item 23-27 DAYS (23 <= days <= 27)
  \item EVERY DAY (days >= 28)
\end{enumerate}

On the other hand, in order to answer the scale-based questions, for each question, we first selected the user's post with the highest computed similarity value. Then, we defined the following intervals to select the scale:

\begin{enumerate}
    \addtocounter{enumi}{-1}
    \item not at all (0 to 0.1)
    \item slightly (0.1 to 0.2)
    \item slightly (0.2 to 0.3)
    \item moderately (0.3 to 0.4)
    \item moderately (0.4 to 0.5)
    \item markedly (0.5 to 0.6)
    \item markedly (0.6 to 1)
\end{enumerate}

This heuristic distinguishes between low and high similarity values so that the higher the similarity, the more likely the response is associated with having the disorder. For instance, if the similarity score computed is 0.65, we chose the answer ``markedly'' for the associated question. 

\subsection{Experimental setup}

As part of our participation in Task 3, three runs have been submitted according to the system developed. They differ in the similarity value between the post and the day-based questions. For \textbf{run 1}, we selected those posts whose similarity is higher than 0.4. For \textbf{run 2}, we established this value at 0.35. Finally, for \textbf{run 3} we increased slightly the value to 0.375. We aimed to observe the difference in the response selected to the questionnaire given by the organizers according to the similarity computed.

Both the experiments in the pre-evaluation and evaluation phases were run on a compute node equipped with a single Tesla-V100 GPU with 32 GB of memory. We used the spaCy sentence transformers (spacy-sentence-bert) library\footnote{\url{https://spacy.io/universe/project/spacy-sentence-bert}} to make use of the transformer model RoBERTa with the default parameters.

\subsection{Results and discussion}
The results obtained with the approaches explored by our team are shown in Table \ref{tab:task1_results3}. The 3 runs have provided similar results, which was to be expected since the only difference between them is in the similarity value established between the posts and the day-based questions. However, the approach that provided the best results was run 2 with the lowest similarity value considered, 0.35. This indicates that perhaps we should relax this value when identifying the set of posts that are related to each question. 

\begin{table}[htp!]
    \caption{Results of SINAI team for Task 3 in ranking-base evaluation}
    \centering
    \begin{tabular}{ccccccccc}
         Run & $MZOE$ & $MAE$ & $MAE_{macro}$ & $GED$ & $RS$ & $ECS$ & $SCS$ & $WCS$\\ 
         \hline
         1 & 0.85 & 2.65 &2.29& 2.63& 3.29 &2.35 & 2.98 &2.40 \\
         2 & 0.87 & 2.60 &2.23& 2.42& 3.01 &2.21 & 2.85 &2.31 \\
         3 & 0.86 & 2.62 &2.22& 2.54& 3.15 &2.32 & 2.93 &2.36 \\
    \end{tabular}
    \label{tab:task1_results3}
\end{table}

In Table \ref{tab:task1_results3_groups}, we show the best run of each participating team and the 3 evaluations given by the organizers assigning all answers to 0 (``all 0''), 6 (``all 6'') and ``average'' which is obtained from the average of the participants' submissions.
\begin{table}[htp!]
    \caption{Results of best run for team for Task 3 in ranking-base evaluation}
    \centering
    \begin{tabular}{ccccccccc}
         Run & $MZOE$ & $MAE$ & $MAE_{macro}$ & $GED$ & $RS$ & $ECS$ & $SCS$ & $WCS$\\ 
         \hline
         IISERB\_2 & 0.92 & 2.18 & 1.76 & 1.74 & 2.00 & 1.73 & 2.03 & 1.92 \\
         SINAI\_2 & 0.87 & 2.60 & 2.23 & 2.42 & 3.01 & 2.21 & 2.85 &2.31 \\
         RELAI\_3 & 0.83 & 3.15 & 2.70 & 3.26& 3.04 & 2.72 & 4.04 &3.61 \\
         \hline
         all 0 & 0.81 & 3.36 & 2.96 & 3.68 & 3.69 & 3.18 & 4.28 &3.82 \\
         all 6 & 0.67 & 2.64 & 3.04 & 3.25 & 3.52 & 3.72 & 2.81 &3.28 \\
         average & 0.88 & 2.72 & 2.22 & 2.69 & 2.76 & 2.20 & 3.35 & 2.85 \\
    \end{tabular}
    \label{tab:task1_results3_groups}
\end{table}

The MZOE measure reflects whether the answers given by the system to complete the questionnaire were correct or not. The closer its value is to 1, the higher the fraction of incorrect predictions. Overall, we can see that the systems presented by the 3 teams make more than 80\% of incorrect predictions, which reflects the difficulty of the task. In our case, the 3 runs provide a similar score, but run 1, based on the similarity heuristic with the highest threshold, 0.4, achieves the higher number of correct answers. However, the MAE and MAE$_{macro}$ measures (range from 0 to infinity, the lower the better) indicate that the responses given by our system are not very far from the real responses, specially those from the run 2. The RS, ECS, SCS, and WCS scores allow us to identify the questions in which our system failed the most, being the most difficult to predict those related to food restriction (questions 1, 2, 3, 4, and 5) and those concerning shape (questions 6, 8, 23, 10, 26, 27, 28, and 11). If we compare our best run with the average results provided by the organisers, we can see that we have a better record except for the RS score, which is above average, and the ECS score which is similar to the average.

In past editions (2018 and 2019) \cite{losada2018overview,davide2019overview}, tasks related to anorexia detection have been presented where the challenge consists of sequentially processing pieces of evidence and detecting early traces of anorexia as soon as possible. However, this year, although the disorder to be focused is the same (anorexia), the formulation of the task is different, being the first time that the challenge aimed at developing an automatic system to fill a standard eating disorder questionnaire based on the evidence found in the user's history. It is worth noting that although the evaluation measures proposed this year are different, this task presents a greater challenge compared to the past editions where the maximum value achieved in terms of F1 score was .71 \cite{davide2019overview}. In addition, this is also reflected in the low number of teams that this task has attracted this year (3 compared to 13 teams in 2019).

\section{Conclusions and future work}

This paper describes our participation as SINAI team in Task 1 and Task 3 of the eRisk@CLEF 2022 edition. The former is the continuation of the first edition in 2021 and aims to detect signs of pathological gambling as soon as possible, while the latter is a new task that focused on measuring the severity of eating disorders signs. For Task 1, we have developed regression and classification models using state-of-the-art pre-trained language models based on Transformers. Besides, for the classification model, we explored a variety of linguistic features including volumetry, lexical diversity, complexity, and emotion detection, achieving the second position among the participants with this model. For Task 3, as no training data was provided by the organizers, we decided to rely on text similarity estimation using word embeddings from Transformers and designing our heuristics. The results achieved in this task as well as the low participation show the difficulty of addressing this type of problem with an automatic system. This fact demonstrates the need to continue investing efforts in this important task.

In future work, we plan to analyze in depth the linguistic features considered in Task 1 to understand to what extent they contribute to the detection of signs of pathological gambling along with further data pre-processing. For Task 3, we would like to perform an error analysis to identify the main weaknesses of our system, as well as explore other Natural Language Processing models.

\begin{acknowledgments}
  This work has been partially supported by Big Hug project (P20\_00956, PAIDI 2020) and WeLee project (1380939, FEDER Andalucía 2014-2020) funded by the Andalusian Regional Government, and LIVING-LANG project (RTI2018-094653-B-C21) funded by MCIN/AEI/10.13039/501100011033 and by ERDF A way of making Europe. Salud María Jiménez-Zafra has been partially supported by a grant from Fondo Social Europeo and Administración de la Junta de Andalucía (DOC\_01073).
  Flor Miriam Plaza-del-Arco has been partially supported by a grant from the Ministry of Science, Innovation and Universities of the Spanish Government (FPI-PRE2019-089310).
\end{acknowledgments}

\bibliography{sample-ceur}




\end{document}